# FAST METHODS FOR RECOVERING SPARSE PARAMETERS IN LINEAR LOW RANK MODELS

Ashkan Esmaeili, Arash Amini, and Farokh Marvasti
Advanced Communications Research Institute (ACRI)
Department of Electrical Engineering, Sharif University of Technology
Tehran, Iran
aesmaili@stanford.edu

## ABSTRACT

In this paper, we investigate the recovery of a sparse weight vector (parameters vector) from a set of noisy linear combinations. However, only partial information about the matrix representing the linear combinations is available. Assuming a low-rank structure for the matrix, one natural solution would be to first apply a matrix completion to the data, and then to solve the resulting compressed sensing problem. In big data applications such as massive MIMO and medical data, the matrix completion step imposes a huge computational burden. Here, we propose to reduce the computational cost of the completion task by ignoring the columns corresponding to zero elements in the sparse vector. To this end, we employ a technique to initially approximate the support of the sparse vector. We further propose to unify the partial matrix completion and sparse vector recovery into an augmented four-step problem. Simulation results reveal that the augmented approach achieves the best performance, while both proposed methods outperform the natural two-step technique with substantially less computational requirements.

*Index Terms*— IMAT; sparse; Lasso; matrix completion; missing data

## 1. INTRODUCTION

The most common approach in dealing with low-rank models containing missing information is to apply matrix completion methods. Matrix completion has been applied to recommendation problems. There are efficient matrix completion methods for low rank models in the literature such as Singular Value Thresholding (SVT) introduced by Candes et al in [1], and Optspace method by Keshavan et al in [2]. In our paper, we focus on Soft-Impute (SI) completion method which was first brought up by Hastie et al in [3]; we are assuming the low-rank model for data and sparsity for the parameters. In [4], Goldberg et al introduced the concept of direct method in recovering the parameters assuming the low-rank structure for the data matrix which consists of missing entries. The parameters vector is not assumed to be sparse in the problem model in [4]. However, in our paper we assume that it is sparse and linked to the compressed sensing problem. Therefore, the problem includes both the compressed sensing and matrix completion problems. The straightforward approach in dealing with missing data in the literature is to apply matrix completion methods to the data at first to learn the missing information. Afterwards, sparse recovery methods could be applied on the learned data to detect the sparse parameters vector. In section 2, we will elaborate upon the new method we propose in dealing with the aforementioned problem. The main concern in working with big data is that we want to avoid time-consuming algorithms. In big data scenarios, completion methods in the literature are time-consuming since they work based on performing Singular Value Decomposition (SVD) which is considered to be of high time-complexity. Therefore, time-efficient methods for this problem are of great importance due to the numerous applications big data is accompanied with.

## 2. PROBLEM MODEL

We consider the problem of finding the sparse signal $\beta$ in the following true linear model:

$$Y = X\beta + \epsilon \qquad (1)$$

where, $X \in R^{m \times n}$ is the data matrix, $\beta \in R^n$ is the parameters signal, $\epsilon \sim N(0, I_{n \times n})$ is the i.i.d noise, and $Y \in R^n$ is the observed labels. We assume $X$ is of low rank which means that $rank(X) = r$ where $r \ll \min(m, n)$.
We also consider $\beta$ to be sparse meaning that the nonzero number of elements in $\beta$ is $s \ll n$.
The support of $\beta$ is defined as follows:

$$Supp(\beta) = \{i \in \{1, \dots, n\} : \beta(i) \neq 0\} \qquad (2)$$

Therefore $|Supp(\beta)| = s$. We also suppose that $X$ has missing entries. For example, we can assume $X$ is generated from an initial $\widetilde{X}$ as follows:

$$= \widetilde{X} \odot B, \text{where } B_{i,j} \sim Ber(\alpha) \qquad (3)$$

where $\odot$ denotes the Hadamard product, and $Ber(\alpha)$ denotes Bernoulli distribution with parameter $\alpha$. In the problem model, we know the matrix $X$ and the labels $Y$. The idea is to use an initial non-complex completion method, then we argue that by applying Iterative Method of Adaptive Thresholding for Compressive Sensing (IMATCS) as in [5], we can find a good initial approximation of the support of the signal $\beta$. Our previous works have shown that IMATCS has good performance in sparse recovery without applying any initial completion on the raw data or with a simple precompletion step; specifically accessing the raw data with missing samples we can have a better estimation of the support by IMATCS rather than LASSO [8]. Afterwards, we proceed considering the raw data on the support columns of the data with missing entries. Then, after dimension reduction, we have less time complexity in recovering the data on support columns. Even if we use an accurate method on the initial data, we have saved large amount of time due to dimension reduction. In other words, we focus on the support of our parameters, and try to be accurate on recovery on the coulmns relating to the support elements. This way, we avoid the time-consuming matrix completion on the entire data. We will include comparisons between the root mean square error (RMSE) values achieved by applying a matrix completion on the entire data followed by sparse recovery and the RMSE values achieved by the introduced approach. To illustrate our approach, we provide the following flowchart for more clarification (coined as the four-step method):

$$X \xrightarrow{MC1} X_{MC1} \xrightarrow{IMAT(\lambda_1)} X_S \xrightarrow{MC2} X_{MC2S} \xrightarrow[IMAT(\lambda_2)]{lasso(\lambda_2)} \hat{\beta}(\lambda_1, \lambda_2)), \quad (4)$$

where $MC1$ denotes the simple matrix completion, $X_{MC1}$ is the data after initial matrix completion. $X_S$ is the data on the support of sparse recovered signal. $X_{MC2S}$ is the reduced data on which accurate matrix completion is applied. $\hat{\beta}$ is the final recovered signal.
Below is the flowchart for the two-step method:

$$X \xrightarrow{MC2} X_{MC2} \xrightarrow[IMAT(\lambda_2)]{lasso(\lambda_1)} \hat{\beta}(\lambda_1) \quad (5)$$

Where $X_{MC2}$ is the entire data completed with accurate completion method, and $\hat{\beta}(\lambda_1)$ is the recovered sparse signal in two-step method. It is worth noting that according to [8], IMATCS outperforms LASSO in recovering the sparse signal parameters and therefore, we have included the results of the IMATCS in the final step of the four-step method and provide comparison with LASSO in sparse recovery at the final step to show how IMATCS yields better RMSE.
The SI method solves the following problem iteratively assuming the data is low-rank.

$$X^*(\lambda) = \text{argmin}\left( \left\| P_E(X - \hat{X}) \right\|_2^2 + \lambda \|X\|_* \right), \quad (6)$$

where $\|X\|_*$ denotes the trace norm of matrix $X$, and $P_E$ denotes the projection on the observed entries of $X$.
In LASSO, we solve the following problem. By cross-validating over $\lambda$ and picking the desired $\lambda$, the sparse signal is recovered.

$$\beta^*(\lambda_1) = \min_{\beta} \left\| \hat{X}\beta - Y \right\|^2 + \lambda_1 \|\beta\|_1 \quad (7)$$

IMATCS method is introduced in [5] and one could refer to this paper to see how it iteratively works in order to find the solution to a compressive sensing problem.
Table 1 best summarizes the four-step sparse signal recovery algorithm. The reason we use SI is that this method works iteratively and we are free to set the trade-off between the accuracy and the complexity of completion. The method we introduced and the original approach are two marginal endpoints of the spectrum which our sensitivity on the number of iterations cover. Our findings and simulations have verified that we could achieve small and even better errors by reducing the complexity using four-step method.

Algorithm 1  Stepwise presentation of four-phase algorithm

**Algorithm 1** Four-step Recovery
1: **Input:**
2: $\mathbf{Y_{m \times 1}}$ : The vector containing the Labels
3: $\mathbf{\hat{X}_{m \times n}}$ : The data matrix containing missing entries
4: $\epsilon$ : Stopping criterion
5: $(\alpha, \lambda_1, \lambda_2)$ Algorithm Parameters
6: **Output:**
7: $\beta^*$ : The reconstructed signal
8: **procedure** FOUR-STEP RECOVERY($\mathbf{Y}, \mathbf{\hat{X}}, \epsilon, \alpha, \lambda_1, \lambda_2$)
9:    Initialization: $\mathbf{X_0} \leftarrow \mathbf{0}_{m \times n}, k \leftarrow 0$
10:    $\beta_{0_{n \times 1}} \leftarrow \mathbf{0}_{n \times 1}, \beta_{1_{n \times 1}} \leftarrow \mathbf{\hat{X}}^\dagger \times \mathbf{Y}$
11:    **while** $\|\beta_k - \beta_{k+1}\|_2 > \epsilon$ **do**
12:       fix $\beta_k$ and solve the following:
13:       $\mathbf{X_k} \leftarrow \min_{X_k} \|P_E(\mathbf{X_k} - \mathbf{\hat{X}})\|_2^2 + \lambda_1 \|\mathbf{X_k}\|_*$
14:       fix $\mathbf{X}_k$ and solve the following:
15:       $\beta^*_{k+1}(\lambda_2) \leftarrow \min_{\beta} \|\mathbf{X_k}\beta - Y\|_2^2 + \lambda_2 \|\beta\|_1$
16:       $k \leftarrow k + 1$
17:    **end while**
18:    $S \leftarrow \{i : \beta_i \neq 0\}, s \leftarrow |S|$
19:    Confine $\mathbf{\hat{X}}$ on supporting columns in $S$, and denote it with $\mathbf{\hat{X}_S}$. Choose $\alpha \ll \epsilon$.
20:    Initialization: $\mathbf{X_0} \leftarrow \mathbf{0}_{m \times s}, k \leftarrow 0$
21:    $\beta_{0_{n \times 1}} \leftarrow \mathbf{0}_{s \times 1}, \beta_{1_{n \times 1}} \leftarrow \mathbf{\hat{X}_S}^\dagger \times \mathbf{Y}$
22:    **while** $\|\beta_k - \beta_{k+1}\|_2 > \alpha$ **do**
23:       fix $\beta_k$ and solve the following:
24:       $\mathbf{X_k} \leftarrow \min_{X_k} \|P_E(\mathbf{X_k} - \mathbf{\hat{X}_S})\|_2^2 + \lambda_1 \|\mathbf{X_k}\|_*$
25:       fix $\mathbf{X}_k$ and solve the following:
26:       $\beta^*_{k+1}(\lambda_2) \leftarrow \min_{\beta} \|\mathbf{X_k}\beta - Y\|_2^2 + \lambda_2 \|\beta\|_1$
27:       $k \leftarrow k + 1$
28:    **end while**
29:    Correct the dimension of $\beta$ by innserting 0 in other indices.
30:    **return** $\beta^*$
31: **end procedure**

## 3. AUGMENTED FOUR-STEP METHOD

In order to increase the prediction accuracy as well as saving time, we propose a second algorithm by concatenating the data matrix $X$ with the column $X\beta$. This again, maintains low rank structure specifically for multi label problems. Next, the resulting matrix is taken into account for imputation and the vector $\beta$ is imputed inside the structure of this matrix. The intuition behind this approach is that the labels help recover the structure of data matrix since it contains useful information about the rank of $X$. One can formulate the problem of finding this vector as follows:

$$\min_{\beta, X} \left\| P_E([X\; X\beta] - [\hat{X}\; Y]) \right\|_2^2 + \lambda_1 \|[X\; X\beta]\|_* + \lambda_2 \|\beta\|_1 \quad (10)$$

This problem is generally a non-convex problem because if we denote $[X\; X\beta]$ with $Z$, we have the following relationship for the last column of $Z$:
$$Z_{*,n+1} = X\beta, \quad (11)$$
which makes non-convex. One approach for solving this minimization problem is to apply coordinate descent method which reduces to the two final stages of four-step method method but the structure of the matrix used is different from considering $X$ alone. It is worth noting that that this method is only different from the four-step method in the third and fourth stage. We will minimize a different objective, and finally we will show in the results section that applying the aforementioned method we enhance the performance.

## 4. ANALYSIS OF TWO-PHASE AND FOUR-PHASE METHODS

In [7], authors have established proofs for bounding the difference between the recovered signal and the original signal in noisy scenario. Here, we came up with the idea of applying theorems in [7] to the matrices completed by SI method, and consider the difference between completed matrices and main data rather than the difference between the noisy matrix and the main data in the final steps of two-phase and four-phase methods.

Let $X_k$ and $\beta_k$ denote the solution to the $k$-th iteration of the SI method. If we apply LASSO on $X_k$, we can recover the support of $\beta_k$ with good approximation. Then we see that by using fewer iterations we can find the support of $\beta_k$. This way, we can find the important features in recovering the parameters. We assume after the $k$-th iteration in Soft-Impute method, we have the following definitions:

$$\beta_k = \operatorname*{argmin}_{\|\beta\|_1 \le b_0 \sqrt{s}} \left\{ \|X_k \beta - Y\|^2 + \lambda_k \|\beta\|_1 \right\} \quad (12)$$

$$\equiv \operatorname*{argmin}_{\|\beta\|_1 \le b_0 \sqrt{s}} \left\{ \frac{1}{2} \beta^T X_k^T X_k \beta - Y^T X_k \beta + \lambda_k \|\beta\|_1 \right\} \quad (13)$$

Let $\hat{\Gamma}$ denote $X_k^T X_k$ and $\hat{\gamma} = X_k^T Y$.
In [7], the authors introduce a condition (lower restricted eigenvalue (RE) condition) as follows:

**Lower RE condition**: $\hat{\Gamma}$ satisfies lower RE with curvature $\alpha_1 \ge 0$ and tolerance $\tau(m,n) \ge 0$ if $\forall \theta: \theta^T \hat{\Gamma} \theta \ge \alpha_1 \|\theta\|_2^2 - \tau(m,n) \|\theta\|_1^2$

In our case, $\hat{\Gamma}$ and $\hat{\gamma}$ are surrogates for $X_\infty$ (the final output of Soft-Impute method) and the original $\beta$, respectively.
the two following bounds are proved in [7]:

$$\left\| \hat{\gamma} - X_\infty^T X_\infty \beta \right\|_\infty \le \phi\big(\sigma(X_k - X_\infty)\big) \sqrt{\frac{\log n}{m}}, \quad (14)$$

$$\left\| (\hat{\Gamma} - X_\infty^T X_\infty) \beta \right\|_\infty \le \phi\big(\sigma(X_k - X_\infty)\big) \sqrt{\frac{\log n}{m}}, \quad (15)$$

where $\sigma(X_k - X_\infty)$ denotes the operator norm of the matrix $X_k - X_\infty$, and $\phi$ is a decreasing function of the singular value. From now on, we denote $\sigma(X_k - X_\infty)$ with $\sigma_D(k)$.
Here, we provide an extension to the main result in [7]. If the Lower RE and the two deviation conditions are met, and if there exist ($\alpha_1, \tau$ such that $\sqrt{s}\tau(m,n) \le \min\{\frac{\alpha_1}{128\sqrt{s}}, \phi(\sigma_D(k))\sqrt{\frac{\log n}{m}}\}$), then for any $\beta^*$ with sparsity $s$ there is a universal positive constant $c_0$ such that for any $\|\beta\|_2 \le b_0$, $\beta_k$ satisfies the following bounds:

$$\|\beta_k - \beta\|_2 \le \frac{c_0 \sqrt{s}}{\alpha_1} \max\left\{ \phi(\sigma_D(k)) \sqrt{\frac{\log n}{m}}, \lambda_k \right\} \quad (16)$$

$$\|\beta_k - \beta\|_1 \le \frac{8 c_0 s}{\alpha_1} \max\left\{ \phi(\sigma_D(k)) \sqrt{\frac{\log n}{m}}, \lambda_k \right\} \quad (17)$$

Now we show how the deviation bounds could be met.

$$\begin{aligned}
\|\hat{\gamma} - X_\infty^T X_\infty \beta\|_\infty &= \|(X_k - X_\infty)^T Y\|_\infty \\
&\le \|(X_k - X_\infty)^T Y\|_2 \\
&\le \|X_k - X_\infty\|_2 \|Y\|_2 = \sigma_{D(k)} \|Y\|_2, \quad (18)
\end{aligned}$$

where the last inequality follows from Cauchy Schwartz inequality. To achieve this bound for this theorem we can set

$$\phi(\sigma_D(k)) = \frac{\|Y\|_2 \sigma_{D(k)}}{\sqrt{\frac{\log n}{m}}} \quad (19)$$

Let $D = X_k - X_\infty$, then,
$\|(X_k^T X_k - X_\infty^T X_\infty)\beta\|_\infty$
$$\le \|X_\infty^T D + D^T X + D^T D\|_2 \|\beta\|_2$$
$$\le 3\sigma_D (2\sigma_\infty + \sigma_D) \|\beta\|_2 \quad (20)$$

In order for the bounds to be valid, we can let $\phi(\sigma_{D(k)}) = \frac{3\sigma_D (2\sigma_k - \sigma_D) b_0}{\sqrt{\frac{\log(n)}{m}}}$.

We can set $\phi(\sigma_{D(k)})$ to be the point-wise maximum of $\left( \frac{3\sigma_D (2\sigma_k - \sigma_D) b_0}{\sqrt{\frac{\log n}{m}}}, \frac{\|y\|_2 \sigma_{D(k)}}{\sqrt{\frac{\log n}{m}}} \right)$ to guarantee that the theorem

holds. So now, what is important for bounding $\|\beta_k - \beta\|_2$ according to the theorem is to make $\sigma_D(k)$ smaller and smaller, to make $\phi(\sigma_{D(k)})$ smaller. What we are looking for is the decrease rate in $\sigma_{D(k)}$. This is controlled in SI algorithm, and that is the reason we pick this up as a completion method. Since the $\sigma_D(k)$ is decreasing in each iteration of SI method, the bound gets tighter and the recovered $\beta$ is guaranteed to converge to the original $\beta$ in both two-step and one four-step methods.

## 5. SIMULATION RESULTS

In this part, we illustrate the results of the simulations. Table 2 shows the runtime required for three algorithms on diverse data sizes. The data is generated by forming the orthonormal matrices and the diagonal matrix containing the singular values as Gaussian random matrices. We observe that the runtime for four-phase and augmented four-phase methods are far less than the time required for the two-phase approach. Table 3 provides the RMSEs for the three methods. As a general rule, the RMSE for four-step method is less than the two step method or in some cases slightly less than that (approximately close). The RMSE for augmented four-phase is generally less than four phase; however, the more runtime is required due to enhanced size of the problem. It is worth mentioning that the runtimes are obtained on a 2.7 GHz Intel Core i7 processor. Fig. 1 shows the cross-validated RMSE over parameters and the comparison between the RMSEs are provided. As we observe the RMSE for the augmented four-step method is less than that of four-step method and the two-step method, respectively. The data is generated by forming the svd and the orthonormal matrices and singular values are assumed to be generated as Gaussian random matrices. The parameters is 15-sparse. As we observe the Augmented four-step method and four-step method outperform the ordinary method in addition to saving more time. In the final stage in the four-step and augmented four-step paper, we also applied the IMAT method in sparse recovery. The RMSEs were quite similar to the case where we applied Lasso in the last step verifying the result in [8]. Thus, we do not provide further tables for the results of IMAT in the last step. We have applied completion and learning algorithm on four-fifth of the data (training data), and found the RMSEs on the test data. The general observation is that the four-phase method outperforms the two-phase and its accuracy could be enhanced by the concept of augmentation as explained.

## 6. CONCLUSION

In conclusion, we notice that large runtime is saved if we restrict the completion on the support of the initial approximation of the parameters vector without losing the performance in the prediction. In order to have an initial approximation of the parameters, we have seen that the IMAT method functions well in sparse recovery. We have found that the four-step method of initial completion followed by applying IMAT (initial sparse recovery), accurate matrix completion on reduced data, and a final sparse recovery is more efficient than the two step method of sparse recovery on the entire data followed by sparse recovery (LASSO) both in terms of the RMSE of prediction on the test set and more importantly computational efficiency. We also improved our method and called it augmented four-step method. It was observed that this method works better in terms of RMSE in comparison to the four-step method while maintaining the same (or slightly little more) amount of time complexity.

TABLE 2
Comparison between the runtimes in seconds achieved by two step, four-step, and Augmented four-step Methods

| Method \ Data size | Two step method | Four step method | Augmented Four step method |
|---|---|---|---|
| m=500, n=200 | 0.9007 | 0.3333 | 0.5518 |
| m=2000,n=200 | 1.0312 | 0.4674 | 1.0072 |
| m=2000, n=500 | 8.4290 | 1.1630 | 1.1300 |
| m=1000,n=200 | 1.1374 | 0.4020 | 0.6493 |
| m=3000,n=500 | 8.9413 | 2.3973 | 2.5233 |

TABLE 3
Comparison between the RMSEs achieved by two-step, four-step, and Augmented four-step Methods

| Method \ Data size | Two step method | Four step method | Augmented four step method |
|---|---|---|---|
| m=500, n=200 | 13.0507 | 18.9203 | 13.6528 |
| m=2000,n=500 | 1.7606 | 1.9290 | 1.7555 |
| m=2000,n=200 | 1.7599 | 1.6499 | 1.6648 |
| m=1000, n=200 | 4.2706 | 3.8700 | 4.2629 |
| m=3000,n=500 | 1.3549 | 1.3614 | 1.2998 |

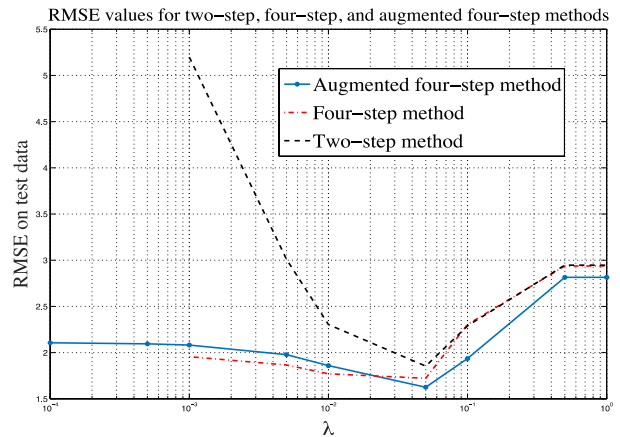

Fig. 1 RMSE values after cross-validation for the three methods on then data with size 2000×500 and 50% missing data and the rank is 100.

# 7. REFERENCES

[1]  Cai, Jian-Feng, Emmanuel J. Candès, and Zuowei Shen. "A singular value thresholding algorithm for matrix completion." *SIAM Journal on Optimization* pp. 1956-1982, 20 Apr (2010).

[2]  Keshavan, Raghunandan H., Sewoong Oh, and Andrea Montanari. "Matrix completion from a few entries." In *2009 IEEE International Symposium on Information Theory*, pp. 324-328. IEEE, 2009.

[3]  Mazumder, Rahul, Trevor Hastie, and Robert Tibshirani. "Spectral regularization algorithms for learning large incomplete matrices." *Journal of machine learning research* pp. 2287-2322., 11 Aug (2010).

[4] Goldberg, Andrew, Ben Recht, Junming Xu, Robert Nowak, and Xiaojin Zhu. "Transduction with matrix completion: Three birds with one stone." In*Advances in neural information processing systems*, pp. 757-765., 2010.

[5]  Azghani, Masoumeh, and Farokh Marvasti. "Sparse Signal Processing." *New Perspectives on Approximation and Sampling Theory*. Springer International Publishing, 2014. 189-213.

[6]  Marvasti, Farokh, Arash Amini, Farzan Haddadi, Mahdi Soltanolkotabi, Babak Hossein Khalaj, Akram Aldroubi, Saeid Sanei, and Janathon Chambers. "A unified approach to sparse signal processing." *EURASIP journal on advances in signal processing* 2012 Feb 22, no. 1 (2012): 1.

[7] Raskutti, Garvesh, Martin J. Wainwright, and Bin Yu. "Minimax rates of estimation for high-dimensional linear regression over-balls." *Information Theory, IEEE Transactions on* 57. pp. 6976-6994, Oct 2011.

[8] A.Esmaeili, F.Marvasti, "Comparison of Several Sparse Recovery Methods for Low Rank Matrices with Random Samples.", arXiv:1606.03672.